\documentclass[11pt,a4paper]{article}
\usepackage[hyperref]{acl2021}
\usepackage{times}
\usepackage{latexsym}
\usepackage{multirow}

\usepackage{graphicx}
\usepackage{amsmath}

\usepackage{amssymb}
\usepackage[normalem]{ulem}
\usepackage{amsthm}
\usepackage{color}
\usepackage{xspace}
\usepackage{subfigure}
\usepackage{booktabs}
\usepackage{array,tabularx}
\usepackage[ruled,vlined]{algorithm2e}
\usepackage{algpseudocode}
\usepackage{amsfonts}

\usepackage{xspace}
\usepackage{microtype}
\usepackage[ruled,vlined]{algorithm2e}
\usepackage{algpseudocode}
\usepackage{microtype}

\aclfinalcopy 

\title{Improving Neural Model Performance through Natural Language Feedback on Their Explanations}

\author{Aman Madaan~\thanks{\hspace{0.5em} authors contributed equally to this work. Ordering determined by dice rolling.}\hspace{0.5em}, Niket Tandon~\footnotemark[1]\hspace{0.35em}$^\dagger$\hspace{0.5em}, Dheeraj Rajagopal~\footnotemark[1]\hspace{0.5em},  Yiming Yang,\\ 
\textbf{Peter Clark$^\dagger$}, \textbf{Keisuke Sakaguchi$^\dagger$}, \textbf{Eduard Hovy} \\
  Language Technologies Institute, Carnegie Mellon University, Pittsburgh, PA, USA \\ 
  $^\dagger$ Allen Institute for Artificial Intelligence, Seattle, WA, USA \\ 
  \texttt{\{dheeraj,amadaan,yiming,hovy\}@cs.cmu.edu} \\ \texttt{\{nikett, peterc,keisukes\}@allenai.org} \\}

\date{}
\definecolor{Red}{rgb}{1,0,0}
\definecolor{Green}{rgb}{0.4,1,0.2}
\definecolor{Blue}{rgb}{0,0,1}
\definecolor{Red}{rgb}{0.9,0,0}
\definecolor{Orange}{rgb}{1,0.5,0}
\definecolor{yellow}{rgb}{0.65,0.6,0}
\definecolor{cadmiumgreen}{rgb}{0.2, 0.7, 0.24}

\newcommand{\red}[1]{\textcolor{Red}{#1}}
\newcommand{\green}[1]{\textcolor{cadmiumgreen}{#1}}

\newcommand{\X}[1]{\mathbf{\mathrm{S}}}
\newcommand{\Z}[1]{\mathbf{\mathrm{C^-}}}
\newcommand{\VV}[1]{\mathbf{\mathrm{C^+}}}
\newcommand{\W}[1]{\mathbf{\mathrm{M^-}}}
\newcommand{\U}[1]{\mathbf{\mathrm{S^-}}}
\newcommand{\Y}[1]{\mathbf{\mathrm{M^+}}}
\newcommand{\LL}[1]{\mathbf{\mathrm{H^-}}}
\newcommand{\M}[1]{\mathbf{\mathrm{H^+}}}
\newcommand{\proscript}[1]{ProScript}

\newcommand{\wiqa}{\textsc{wiqa}\xspace}

\newcommand{\ours}{\textsc{mercurie}\xspace}  

\newcommand{\calM}{$\mathcal{M}$\xspace}

\newcommand{\calG}{$\mathcal{G}$\xspace}
\newcommand{\GENS}{$\mathcal{M}^{*}$\xspace}
\newcommand{\GEN}{\calM}
\newcommand{\CORR}{\calG}

\newcommand{\upd}{$\mathbf{S}$\xspace}
\newcommand{\hypo}{$\mathbf{H}$\xspace}
\newcommand{\pre}{$\mathbf{P}$\xspace}
\newcommand{\phu}{$\mathbf{PHS}$\xspace}

\newcommand{\utype}{\textbf{T}\xspace}

\newcommand{\atomic}{\textsc{atomic}\xspace}
\newcommand{\snli}{\textsc{snli}\xspace}
\newcommand{\social}{\textsc{social}\xspace}

\def\@withdot.{\ifmmode\!\string/\!
               \else\kern-1.8pt\string/\kern-1.8pt\fi.}

\newcommand{\nle}{\textsc{nl-edit}\xspace}

\newcommand{\squishlist}{
  \begin{list}{$\bullet$}
    { \setlength{\itemsep}{0pt}      \setlength{\parsep}{3pt}
      \setlength{\topsep}{3pt}       \setlength{\partopsep}{0pt}
      \setlength{\leftmargin}{1.5em} \setlength{\labelwidth}{1em}
      \setlength{\labelsep}{0.5em} } }
\newcommand{\reallysquishlist}{
  \begin{list}{$\bullet$}
    { \setlength{\itemsep}{0pt}    \setlength{\parsep}{0pt}
      \setlength{\topsep}{0pt}     \setlength{\partopsep}{0pt}
      \setlength{\leftmargin}{0.2em} \setlength{\labelwidth}{0.2em}
      \setlength{\labelsep}{0.2em} } }

 \newcommand{\squishend}{
     \end{list} 
 }

\begin{document}
\maketitle

\begin{abstract}
A class of explainable NLP models for reasoning tasks support their decisions by generating free-form or structured explanations, but what happens when these supporting structures contain errors? Our goal is to allow users to interactively correct explanation structures through natural language feedback. We introduce \ours - an interactive system that refines its explanations for a given reasoning task by getting human feedback in natural language. Our approach generates graphs that have 40\% fewer inconsistencies as compared with the off-the-shelf system. Further, simply appending the corrected explanation structures to the output leads to a gain of 1.2 points on accuracy on defeasible reasoning across all three domains.\footnote{We release a dataset of over 450k graphs for defeasible reasoning generated by our system at \url{https://tinyurl.com/mercurie}.}
\end{abstract}

\section{Introduction}

\begin{figure*}[!ht]
\centering
{\includegraphics[scale=0.35]{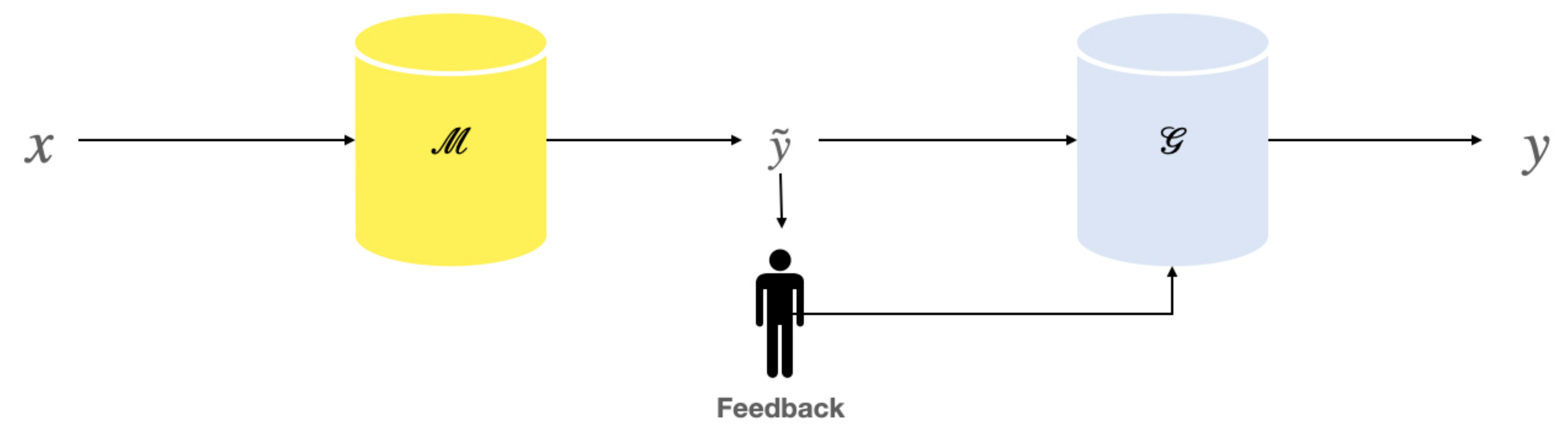}}
\caption{Our pipeline: the output generated by \GEN is corrected by \CORR using human feedback.}
\label{fig:pipeline}
\end{figure*}

Interactive Machine Learning allows humans to give feedback to the models, often leading to improved accuracy \citep{Fails2003InteractiveML,Raghavan2006ActiveLW,Settles2011ClosingTL}. Interactive systems for NLP have used human-in-the-loop style interactions for helping refugee settlement \citep{Brown2016DesigningFT}, aligning topic models \citep{Yuan2018MultilingualAI} and enhancing bilingual word embeddings \citep{Yuan2020InteractiveRO}. Neural models have made advancements in explanation generation but are expensive to retrain.
This paper aims to improve the model output through natural language feedback (e.g., on its explanation) without retraining. 

One line of prior approaches (interactive semantic parsing approach) \cite{elgohary2021nl, percy-2016-learning-language-games-interaction-shouldrn} parse natural language user feedback into a set of edit operations, which can then be executed on the incorrect explanation structure, thereby correcting the explanation.
In these approaches, the feedback is specific to a semantic parsing schema and has to be specialized, i.e., directly mapped to specific instructions or literals, limiting its generalizability. 
Moreover, the feedback is expected to be actionable, containing a specific set of edit operations expressed in natural language. However, real-world human feedback is often imprecise and not directly actionable. Another line of prior approaches (interactive reasoning approach)  \citep{Talmor2020TeachingPM} explore interactivity by enriching the context of an input sample through human feedback. However, for the human giving the feedback, the model is a black box -- so the human does not know what the model's internal belief is and how it will change based on the feedback. 

These two lines of prior approaches inspire this paper -- we provide more transparency to the human than the interactive reasoning approach as the model receives feedback on the explanation (similar to the interactive semantic parsing approach). We do this while relaxing the assumptions of the parsing approach -- our feedback does not have a task-specific structure, and it is not assumed to be actionable (similar to the interactive reasoning approach).

We introduce \ours, a pipeline system with two components, a previously trained neural model $\mathcal{M}$ and a graph corrector $\mathcal{G}$. It takes as input any previously trained neural model $\mathcal{M}$ capable of generating an explanation structure. The second input is a natural language human feedback on the generated explanation structure (for example, that some nodes are inconsistent with the rest of the graph). As output, it produces a better explanation structure.
 

The contributions of this work are:
\squishlist
\item We demonstrate a system that shows that an explainable NLP model output can be improved through natural feedback on their explanations. Experiments show that \ours can improve the consistency of explanation structures by up to 40\% (\S\ref{sec:feedback-defeasible}).
\item We also show downstream task (defeasible inference \citep{rudinger-etal-2020-thinking}) improvement for all domains by at least 1.2 points on accuracy (\S\ref{sec:results}).
\squishend

\SetKwRepeat{Do}{do}{while}
\SetKwInput{KwGiven}{Given}
\SetKwInput{KwInit}{Init}
\SetKwInput{KwTrainCorr}{Train \CORR on $\mathcal{D}_{\mathcal{G}}$}
\SetKwInput{KwInference}{Inference}
\begin{algorithm}[ht]
\SetAlgoLined
\KwGiven{\GEN: $x \rightarrow  \tilde{y}$, $\{x_i\}_{i=1}^{N}$ } 
\vspace{0.7em}
\KwTrainCorr{}
\Indp
$\mathcal{D}_{\mathcal{G}} = \emptyset$\;
\For{$i \gets 1, 2, \ldots, N$}{
$\tilde{y}_i = \mathcal{M}(x_i)$\;
$I_i = \texttt{feedback}(\tilde{y}_i)$\;  %
$y_i = \texttt{human}(\tilde{y}_i)$\; 
$\mathcal{D}_{\mathcal{G}} = \mathcal{D}_{\mathcal{G}} \cup (x_i, I_i, \tilde{y}_i, y_i)$\;
}

\Indm
\Indp Train \CORR on $(x, I, \tilde{y}) \rightarrow  y$ \;  
\Indm
\vspace{0.7em}
\KwInference{}  
\Indp

$\tilde{y} = \mathcal{M}(x)$\;

\While{$I: \texttt{feedback}(\tilde{y}) \neq \emptyset$} {
$\tilde{y} = \mathcal{G}(x, I, \tilde{y})$\;
}
$y = \tilde{y}$\;

\caption{\ours algorithm to correct explanations through human feedback}
\label{alg:alg-1}
\end{algorithm}

\section{Related work}

\paragraph{Interactive Learning:} Interactive learning involves a human in the loop, as opposed to learning from datasets collected offline.
Relevant approaches in NLP are wide-ranging from active learning \cite{Raghavan2006ActiveLW,wu2019active} to training dialogue systems that adapt to user utterances, spanning diverse domains \cite{Holzinger2016InteractiveMLHealth}. 
There are various modes of interaction (through labels \cite{Raghavan2006ActiveLW, Fails2003InteractiveML}, utterance \cite{Radlinski2019DialogToSeqOfQAPairs}, imitation \cite{Brantley2020ActiveImitationlearning}, and language \cite{elgohary2020speak}). Our work uses language as the mode of interaction. 

\paragraph{Language-based interactions:} Natural language interaction allows for expressive human feedback to correct a model. 
In language-based interactions, controlled settings \cite{Mehta2019interactionRobotUsingAdvice, percy-2016-learning-language-games-interaction-shouldrn} give a better handle and are easy to evaluate. However, they do not generalize to real-world settings--  human feedback is rich, and it is not desirable to be restricted to a vocabulary.
Finally, the model being taught is treated either as (i) a black box (as in machine teaching \cite{Dasgupta2019MachineTeaching}, \cite{Talmor2020TeachingPM}) or (ii) the beliefs of the model are in some form exposed to feedback (as in interactive semantic parsing \cite{elgohary2021nl}). This paper is uniquely positioned because we present the first system, which has interaction through language by directly giving feedback on the model's beliefs (explanation) in a real-world, open domain setting. 

\paragraph{Interactive Semantic Parsing:} The common theme in prior approaches to this task based on interactive semantic parsing (such as \cite{elgohary2021nl, percy-2016-learning-language-games-interaction-shouldrn}) is that user feedback is mapped into structure edit commands, which can then be executed on the incorrect structures to fix it. For example, \cite{elgohary2021nl} presented \nle to fix SQL queries using human feedback such as: 
\texttt{replace course id with program id.}. 
However:
\squishlist
\item the feedback are syntactic with a certain task-specific formal structure, e.g., \nle is known to struggle with natural feedback that does not describe an edit directly \cite{elgohary2021nl}.
\item the feedback is expected to be actionable. Rather than highlighting a problem or error, it is expected to contain a solution to fix the error.
This feedback is then parsed using semantic parsing techniques into a set of structure edit commands.
\squishend

\paragraph{Differences w.r.t. Interactive Semantic parsing}
Unlike \nle, we do not make assumptions about the structure of the feedback. Moreover, we assume that the feedback would be non-actionable (pointing out some local or global error without providing a solution to fix the error). This should especially hold with the growing complexity of the structure to give feedback because it is simpler for a human to point to the problem rather than enumerate (in natural language) the edits that might be required. Therefore, semantic parsing techniques do not apply to our problem as the feedback is non-actionable (i.e., our feedback only highlights that something is wrong, not how to fix it).

\paragraph{Interactive learning for reasoning tasks}
Our focus is a reasoning task that accounts for the context and requires commonsense to bridge between the feedback to a possible solution. In this, we are inspired by \cite{Talmor2020TeachingPM} where the interaction is with a black box system (unlike this paper), and when the model incorrectly answers whether \texttt{A whale has a belly button}, then a user tells the model the explicit rule \texttt{A mammal has a belly button}, the model corrects its answer by combining the feedback with its implicit knowledge, e.g., that \texttt{A whale is a mammal}. Our work extends along this line of research by showing that a model can update a model's explanation structure in a reasoning task setting. 

\section{Task and Dataset}
\label{sec:task-defeasible}

We focus on the task of generating graphs for defeasible inference queries. After presenting the task, we describe the graph generator \GEN that generates an inference graph for a defeasible inference query. Subsequently, we will use the feedback described in \S \ref{sec:feedback-defeasible} to train \CORR, a system that fixes the output generated by \GEN.

\subsection{Task: Defeasible Inference}

Defeasible inference~\cite{rudinger-etal-2020-thinking} is a mode of reasoning in which given a premise \pre, a hypothesis \hypo may be strengthened or weakened in light of new evidence.
For example, given a premise \textit{ocean causes erosion}, the hypothesis \textit{rocks become smaller} will be strengthened by the situation \textit{waves are bigger}, and weakened by the situation \upd \textit{no waves}.
We use \phu to refer to a defeasible query and \utype to the answer~(strengthened or weakened). 

This problem has been widely studied in cognitive science by supporting defeasible inference through argumentative frameworks \cite{Pollock1987DefeasibleR}. Humans have found argumentations helpful in defeasible reasoning, and this insight has led to models that simulate argumentations through an \emph{inference graph}, e.g., \citet{Pollock2009ARS} supplement defeasible queries \phu with an \textit{inference graph}. An inference graph contains events as nodes and the causal relationship between the nodes as edges. The motivation behind using inference graphs is to provide additional context for each \phu query that might help the humans understand the nature of the effect that an update situation \upd has on the hypothesis. Being costly to construct by hand, inference graphs have only been studied at a small scale. 

In the absence of a large repository of inference graphs for defeasible queries, we propose their automatic generation by learning from \wiqa~\cite{tandon2019wiqa} - a repository of graphs that are similar to an inference graph ( Section~\ref{sec:inferenceinfluencemapping}). 
The main challenge in learning from these graphs is that they are narrowly focused on the procedural text domain. In contrast, defeasible inference task has a wide scope-- thus requiring the transfer technique that we present in \ref{sec:gen-for-defeasible}.
Given the central role that the \wiqa dataset plays in our work, we provide a brief description next.

\subsubsection{\wiqa}
\begin{figure}[!ht]
\centering
{\includegraphics[scale=0.38]{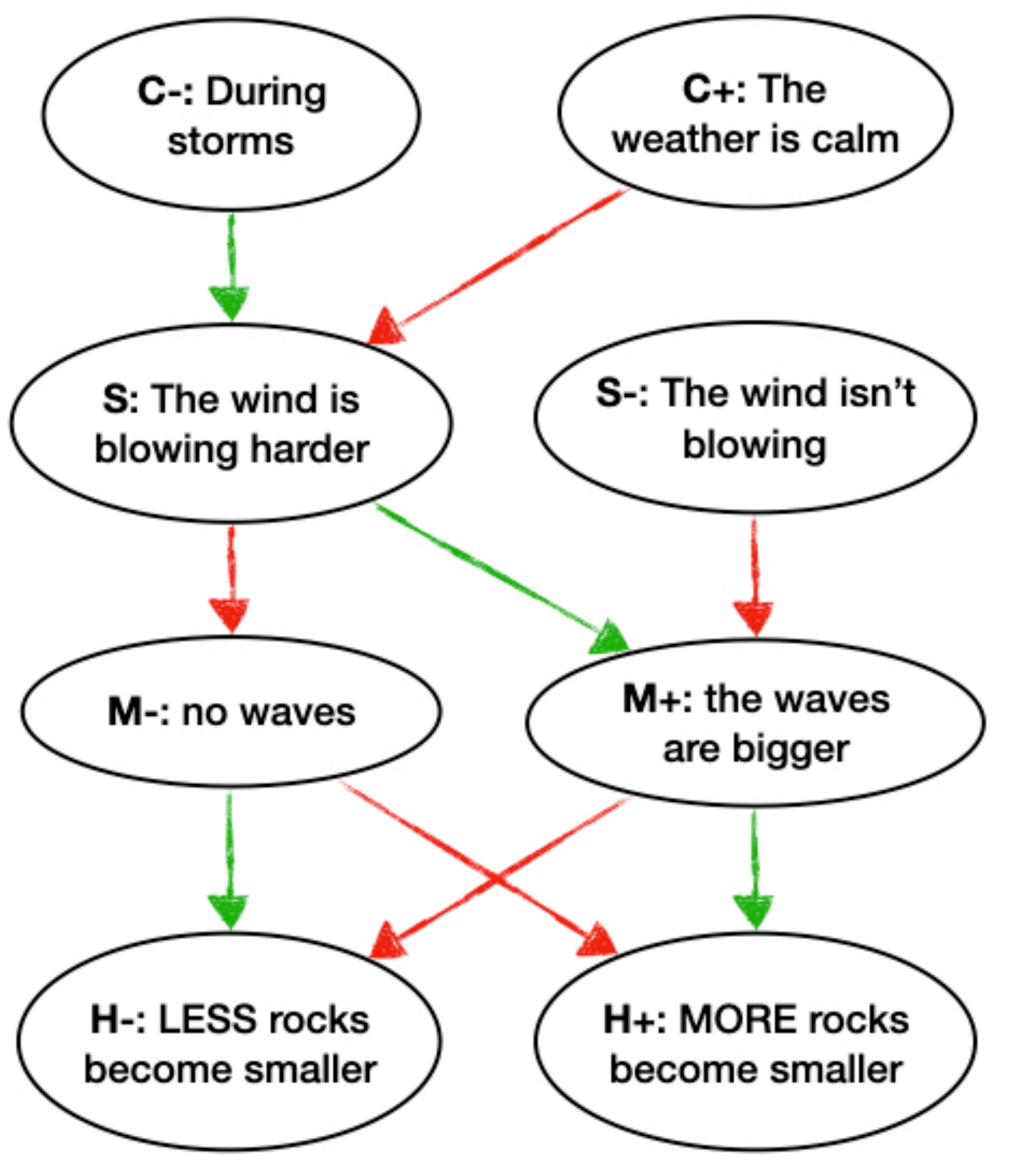}}
\caption{A sample influence graph about spread of a virus during a pandemic}
\label{fig:virusinfluencegraph}
\end{figure}

\wiqa comprises of 2107 pairs of $(P, G)$ where $P$ is a paragraph that describes a process (e.g., the spread of a virus).
The influence graph $G$ corresponding to $P$ is a directed acyclic graph (\textsc{dag}) that captures the interactions between the events and their influences within the context of the process described by $P$. 
Let $G=(V, E)$, where $V$ denotes the set of vertices and $E$ the set of edges.
The nodes $n \in V$ are events relevant to the process.
Each node $n$ is described by a sequence of text tokens.
The edge set $E$ contains two types of edges: $\textit{helps}$ and $\textit{hurts}$, denoted by \green{green} and \red{red} arrows respectively.
A $\textit{helps}$ edge between a source node $n_c$ and a target node $n_e$ signifies that the source event $n_c$ positively influences the target event $n_e$ and a $\textit{hurts}$ edge stands for $n_c$ negatively influencing $n_e$.
Figure~\ref{fig:virusinfluencegraph} shows an example influence graph for the process of ``spread of a virus during a pandemic.''

\subsubsection{\wiqa as a repository of inference graphs}
\label{sec:inferenceinfluencemapping}
We show that the nodes of an influence graph in \wiqa are similar to the inference graph for defeasible reasoning proposed in ~\cite{Pollock2009ARS}, by showing a semantic mapping between the components of a defeasible query and an influence graph.

\squishlist
    \item The premise of a defeasible query \pre and the passage in \wiqa both play a similar role of providing more context for the influence graph.
    \item Each \wiqa graph has two hypothesis nodes, which capture either the strengthening or weakening of a hypothesis.
    Thus, there is a natural correspondence between the hypothesis nodes in \wiqa and the hypothesis in defeasible.
    \item Each influence graph consists of a node $S$, which contains an event grounded in $P$ that signifies a change. This is similar to the update \upd in the defeasible query.
\squishend

\subsection{Designing \GEN for Defeasible Reasoning}
\label{sec:gen-for-defeasible}
\begin{figure*}[!ht]
\centering
{\includegraphics[width=0.90\textwidth,height=0.17\textheight]{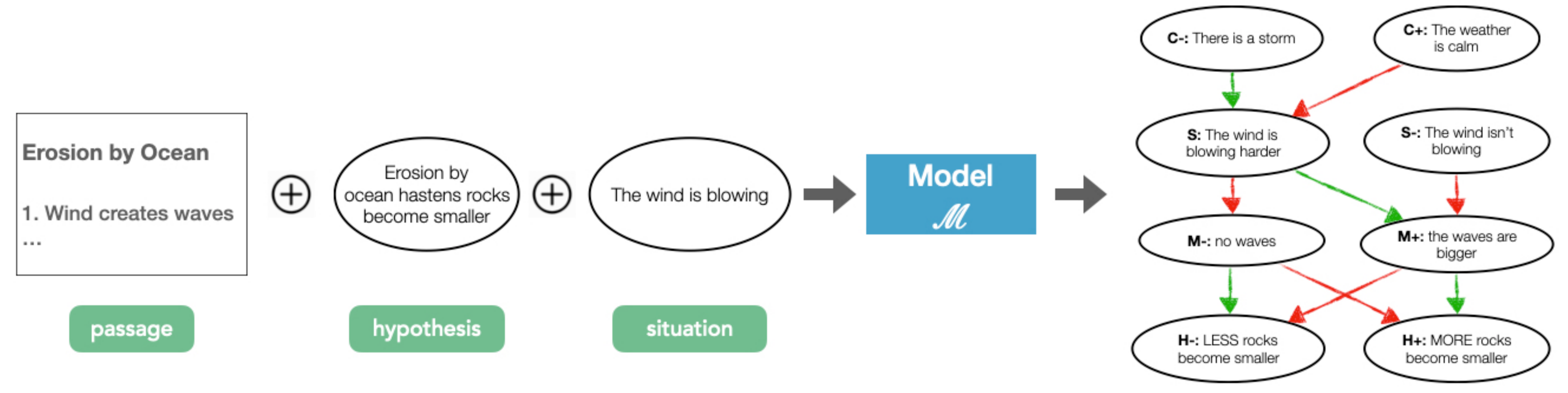}}
\caption{Training the graph generator $\mathcal{M}$ for defeasible reasoning.}
\label{fig:deftrain}
\end{figure*}

Given these similarities, we train a graph-generator on \wiqa and transfer it for defeasible reasoning.
Our goal is to supplement each defeasible query \phu with an inference graph. We first train a graph generator \calM using \wiqa.
As discussed, each example in \wiqa consists of a $(P, G)$ pair, where $P$ is the passage, and $G$ is the influence graph.
We extract the hypothesis node $H$ and the situation node $S$ from $G$ (using the last two nodes in Figure~\ref{fig:virusinfluencegraph}). 
We then train a sequence-to-sequence generation model (based on T5-11B), where the input is the string $P \Vert H \Vert S$ and the output is the corresponding influence graph $G$ encoded as a string.
During inference, we obtain a graph for the defeasible query \phu by setting passage = \pre, hypothesis = \hypo, and situation = \upd, as discussed. Figure~\ref{fig:deftrain} shows the details of the training process.

\section{Human feedback on $\mathcal{M}$}
\label{sec:feedback-defeasible}

In this section, we propose a method to take feedback on the output of \GEN.

\subsection{Human feedback}

We evaluate the graphs produced by \calM for defeasible reasoning using human evaluators. Two human judges evaluated 100 graphs produced by \calM. 
The judges found that all the graphs had the correct structure, but 70\% of them had \textit{repeated} nodes with the same information.

Each node in an influence graph plays a specific role (e.g., positive contextualizer or mediator). Thus, repeated nodes violate the semantic structure of a graph.
Additionally, they also reduce the amount of information carried by each graph.
For defeasible reasoning, we focus on reducing this repetition of nodes in each graph.
We note that we do not utilize the edge structure of the graph for this work or take feedback on it.
The structure of the graphs is assumed to be fixed.
Our intuition is that reducing the number of repeated nodes will improve the quality of these graphs, making them more useful for downstream tasks.
To be consistent across tasks, we refer to such graphs with repeated nodes as being \textit{incorrect graphs}.

\SetKwInput{KwGiven}{Given}
\SetKwInput{KwInit}{Init}
\begin{algorithm}[!h]
\SetAlgoLined
\KwGiven{Inference graphs G generated by \GEN, and G* generated by \GEN*}
\KwResult{Training data for \CORR}
\KwInit{$\mathcal{D} \gets [ ]$}
\For{$i \gets 1, 2, \ldots, |\mathcal{M}|$}{
$F_{G_i} = feedback(G_i)$ \;
$F_{G_i}* = feedback(G_i*)$\;
\uIf{$F_{G_i} \not = \emptyset$ and $F_{G_i}* = \emptyset$}{
\tcc{$G_i$ has problems, $G_i*$ is good}
$\mathcal{D} = \mathcal{D} \cup (G_i, F_{G_i}, G_i*)$\;
} \uElseIf {$F_G = \emptyset$ and $F_G* = \emptyset$} {
\tcc{Both $G_i$ and $G_i*$ are good}
$\mathcal{D} = \mathcal{D} \cup (G_i, \text{No issues, looks good}, G_i*)$\;
}
}
\KwRet{$\mathcal{D}$}
\caption{Generating training data for \CORR using human feedback.}
\label{alg:alg-train-for-corr}
\end{algorithm}

\subsection{Automating human-like feedback}
\label{sec:automate-human-like-feedback}
We observed that humans found it cognitively challenging to look at multiple nodes and check for the consistency of nodes and repetition of content across multiple unrelated or opposite polarity nodes. In contrast, prior work on assembling structure edit commands relies on the relative simplicity of the structure (such as in an SQL query), allowing targeted feedback. This is not possible in our case, owing to the sizeable cognitive load of manually labeling each node while maintaining structural consistency. Therefore, using human annotations, we devised a simple rule-based system $F$ that uses token-based overlap to detect repetitions while preventing spurious matches due to negation. 
Figures \ref{fig:snliexample}, \ref{fig:atomicexample}, and \ref{fig:socialexample} show examples of various kinds of inconsistencies and the corresponding feedback.

\subsection{Automating expected corrected graph}
\label{sec:automate-expected-corrected-graph}
Ideally, it would be desirable to have training data that provides a \textit{fixed graph} corresponding to each incorrect graph. 
However, we realized that manually fixing incorrect graphs is not scalable, as it requires identifying repeated nodes and then coming up with a label that would remove the repetitions across the graph.
We circumvent this issue by training another version of graph generator \GENS.
The training process of \GENS closely follows that of \GEN: we set the input to $P \Vert H \Vert S \Vert T$ and the output to $G$.
Note that the only difference here from \GEN is that the generation is now additionally conditioned on the edges, leading to more diverse and possibly less noisy graphs.

During inference, we obtain a graph for the defeasible query \phu by setting passage = \pre, hypothesis = \hypo, and situation = \upd, as discussed. Figure~\ref{fig:deftrain} shows the details of the training process.

We further note that such conditioning is not possible for the general case since the graph edges are not available with defeasible queries.

We use T5-11B~\cite{raffel2020exploring} as our graph generator \GEN, feedback graph generator \GENS, as well as graph corrector \CORR.
\begin{figure*}
\centering
\begin{minipage}{.5\textwidth}
  \centering
  \includegraphics[width=.80\linewidth]{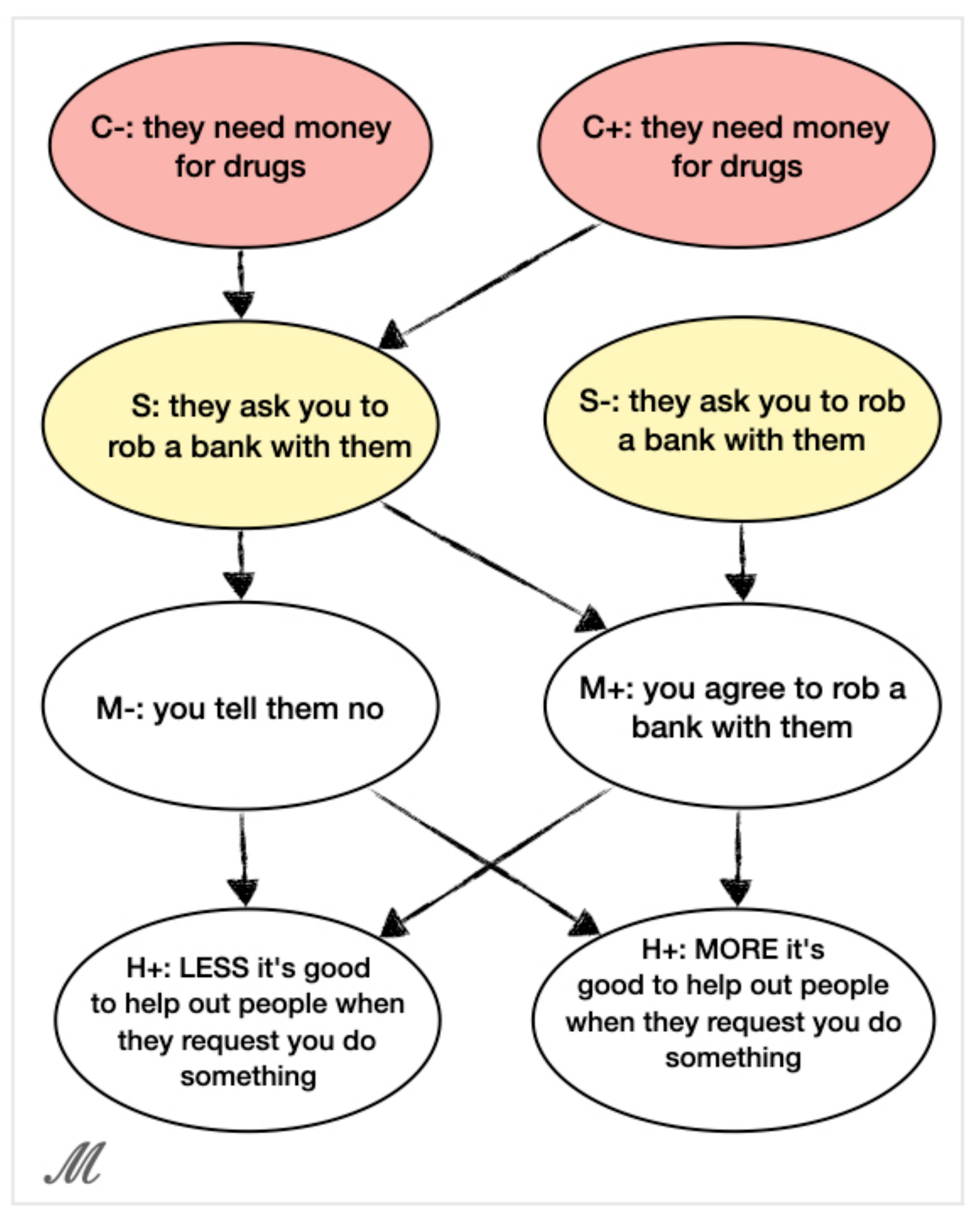}
   \caption{\textit{C-, C+ and S,S- are overlapping.}}
\end{minipage}%
\begin{minipage}{.5\textwidth}
  \centering
  \includegraphics[width=.8\linewidth]{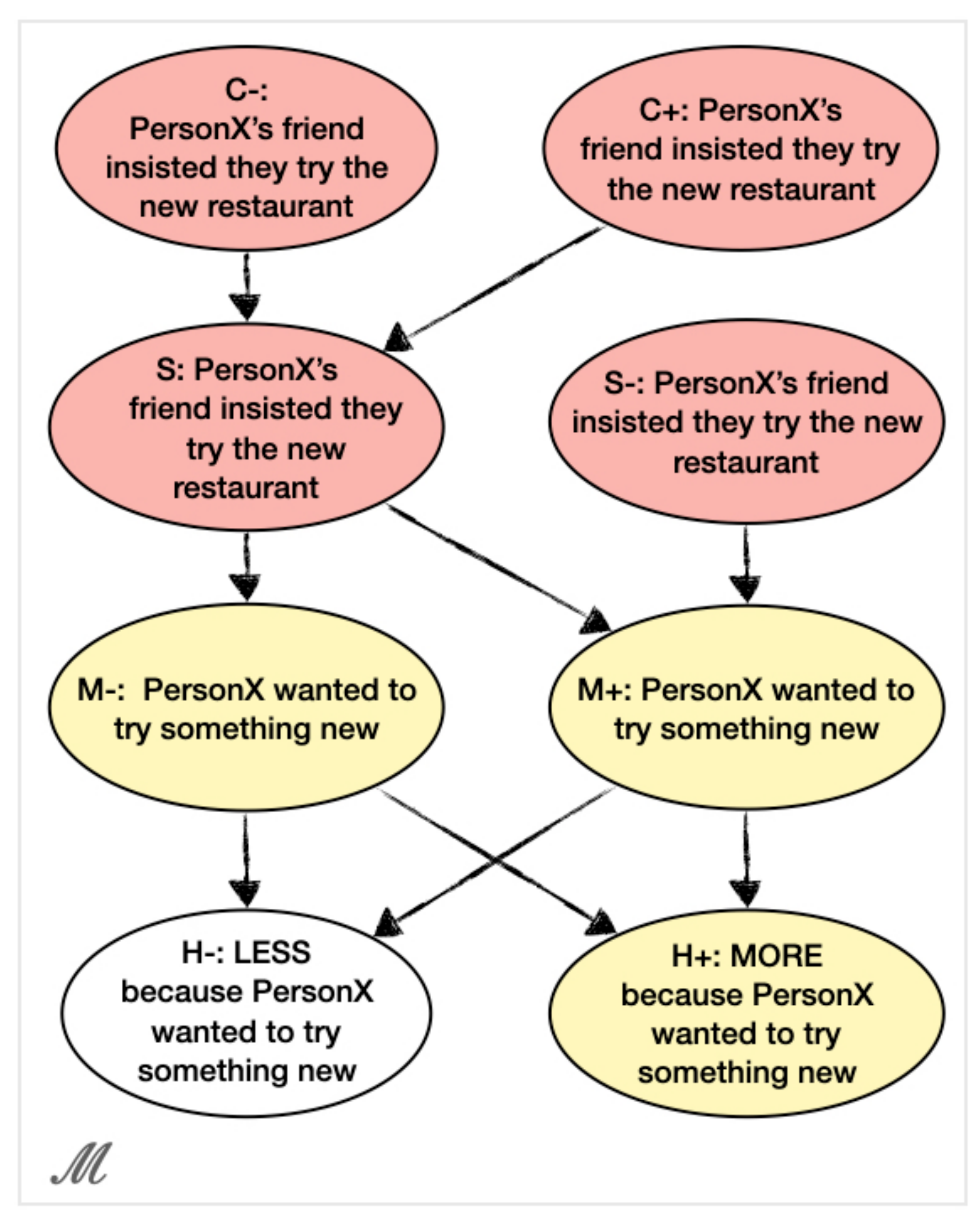}
  \caption{\textit{C-,C+,S,S-  and M-, M+, H+ are overlapping.}}
\end{minipage}
  \captionof{figure}{Incorrect graphs generated by \GEN for SNLI (left) SOCIAL domains of Defeasible. The feedback on each graph is mentioned in caption, and we provide the fixed versions of these graphs in the Appendix.}
  \label{fig:feedbackexample}
\end{figure*}
\section{Correcting explanation structure through human feedback}
\label{sec:rq1}

\begin{figure*}[!htb]
\minipage{0.33\textwidth}
  \includegraphics[width=\linewidth]{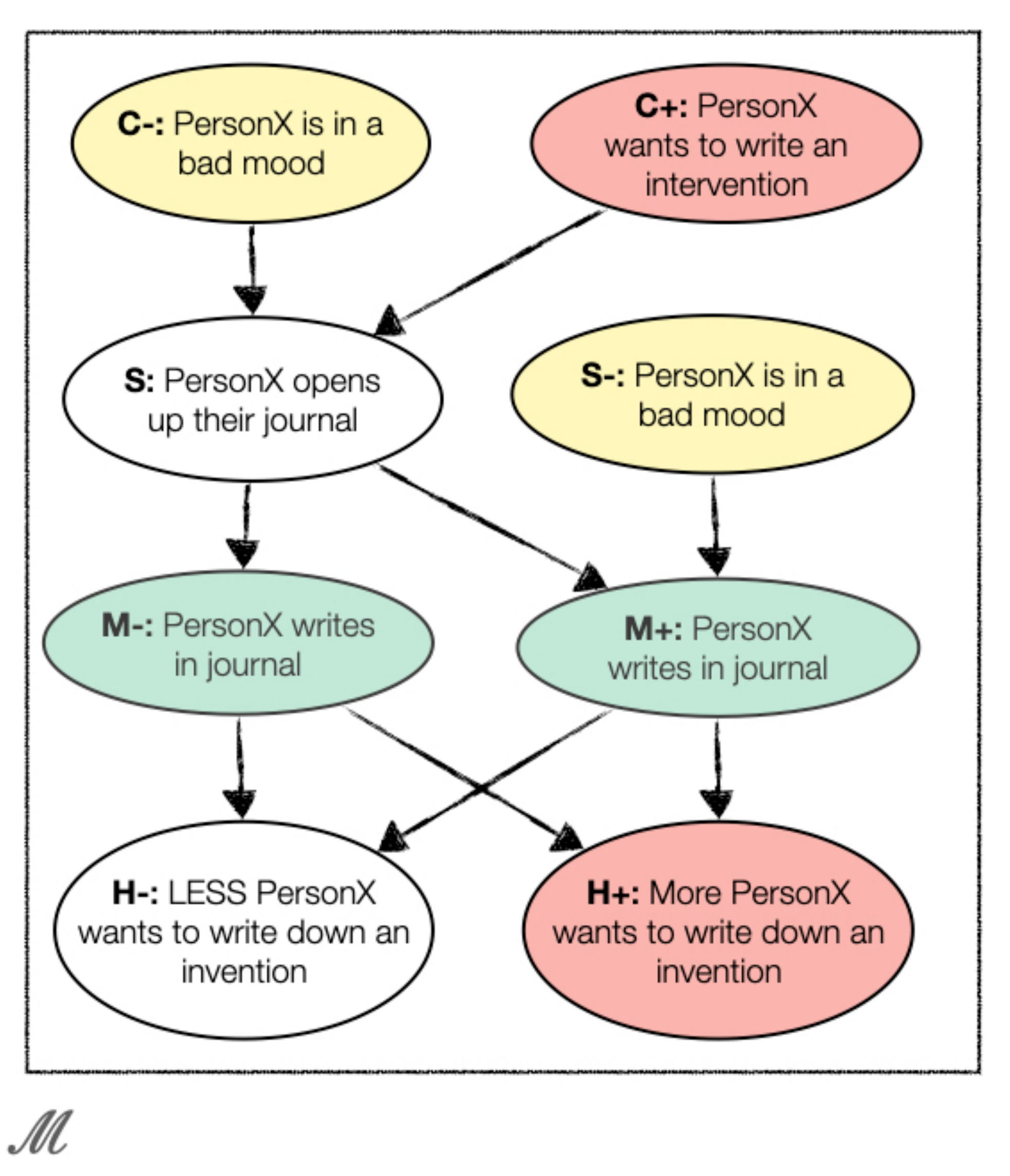}
\endminipage\hfill
\minipage{0.33\textwidth}
  \includegraphics[width=\linewidth]{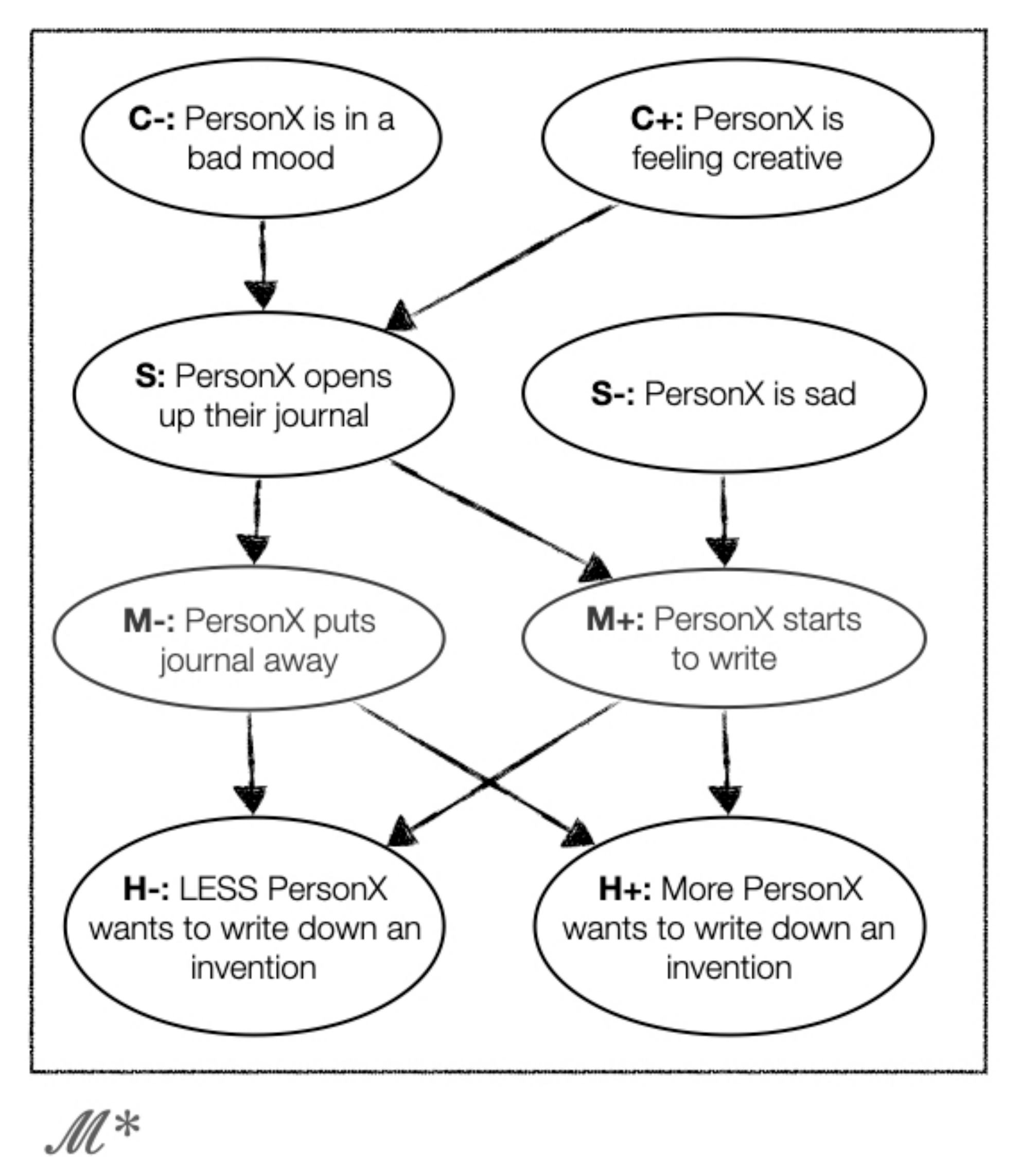}
\endminipage\hfill
\minipage{0.33\textwidth}%
  \includegraphics[width=\linewidth]{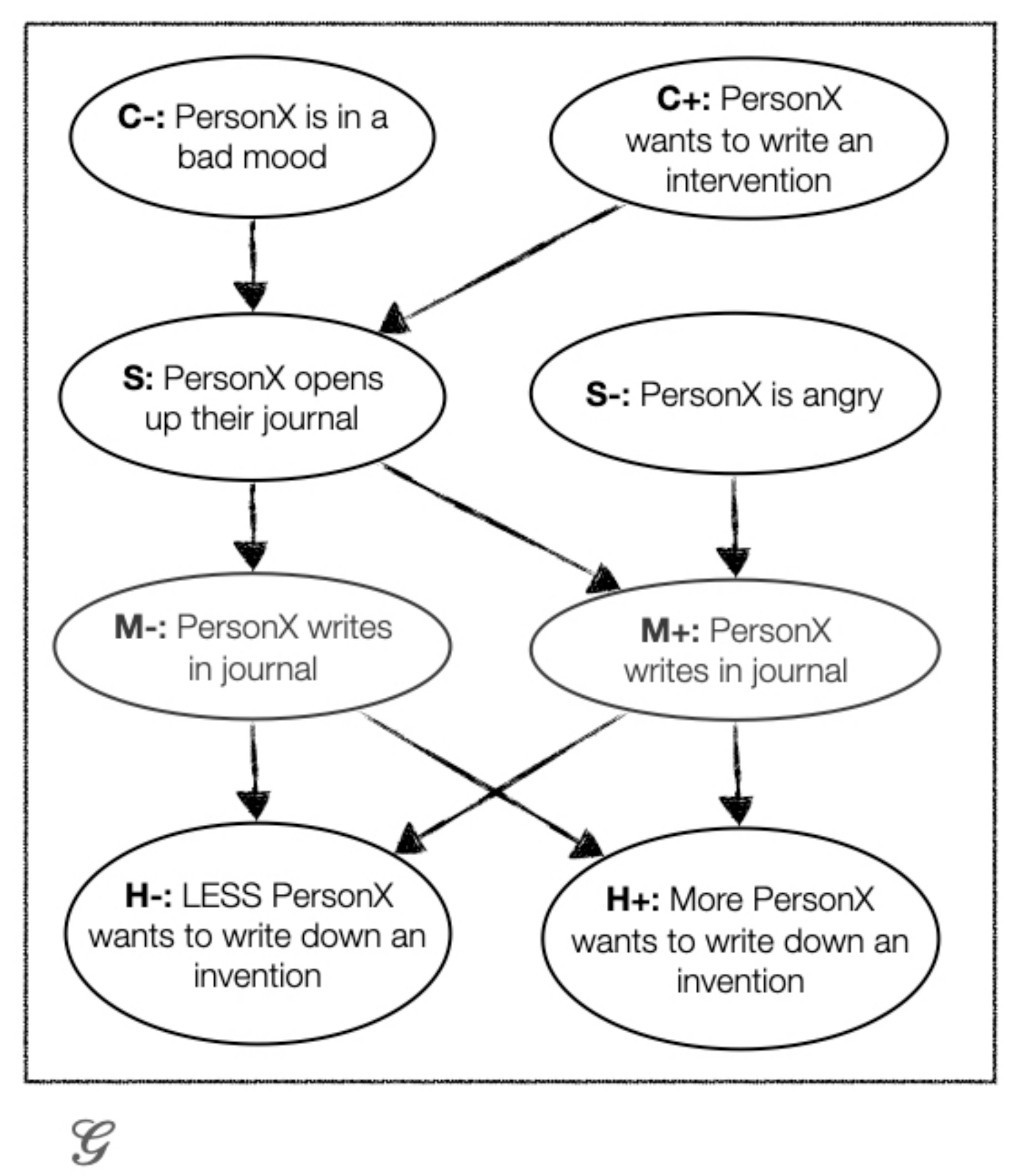}
\endminipage
\caption{The graphs generated by \GEN (left), \GENS (middle), and \CORR (right).The input graph has repetitions for nodes $\{C{-}, S{-}\}$,  $\{C{+}, H{+}\}$, and $\{M{-}, M{+}\}$. The corrected graph replaces the repetitions with meaningful labels.}
\label{fig:defeasible_incorrect}
\end{figure*}

Can we make use of the feedback described \S \ref{sec:feedback-defeasible}? We show that we can train a model, \CORR, that takes that feedback and improves \GEN. That is, given \phu, \GEN generates a potentially noisy graph (\S \ref{sec:gen-for-defeasible}) - and \CORR learns to correct this graph using the automatic human-like feedback (\S \ref{sec:automate-human-like-feedback}) and compute loss over the expected corrected graph (\S \ref{sec:automate-expected-corrected-graph}). First, we show this graph correction system \CORR, followed by empirically measuring the effectiveness of \CORR.

\subsection{Training the graph corrector \CORR}
We now proceed to train the graph corrector \CORR using Algorithm~\ref{alg:alg-train-for-corr}. 
The \CORR is also trained as a sequence-to-sequence model.
For a given query \phu, the graphs generated by \GEN and \GENS are first paired.
From these pairs, we only retain those cases where the \GEN graph is incorrect, whereas \GENS graph is not, as identified by our rule-based feedback system $F$.
We record each such example as $(G', F(G'), G*)$.
We also retain pairs where both $G'$ and $G''$ are correct, and in those cases, the feedback $F(G')$ is set to \textit{no issues, looks good}.
This is then fed to our generation model, which is trained to generate $G^{*}$ from $G', F(G')$.

Training \CORR completes our pipeline for obtaining high-quality graphs for each defeasible query.
First, given a defeasible query \phu, we generate a potentially incorrect graph. 
G' using \GEN.
We then use the feedback generator $F$ to obtain feedback $F(G')$ on $G'$.
The tuple $(G', F(G'))$ is then fed to \CORR to obtain a corrected graph G.

\section{Results}
\label{sec:results}

In this section, we answer two questions: i) Does \CORR reduce the inconsistencies in the graphs? ii) Does using graphs generated by \CORR help the end task?

\subsection{Does \CORR reduce the inconsistencies in the graphs?}

We evaluate the repetitions in the graphs using two metrics:
\squishlist
    \item \textbf{rep. per graph:} the average number of repeated nodes in the graphs produced by \GEN and \CORR.
    \item \textbf{\% with repetitions:} the percentage of graphs with at least one repeated node.
\squishend

As Table~\ref{tab:rq1-main-defeasible} shows, \CORR reduces the average repetitions by 40\% (2.11 to 1.25) and reduces the fraction of graphs with at least one repetition by 25.7 on average.

\begin{table}[!h]
\begin{tabular}{p{0.075\textwidth}p{0.09\textwidth}|p{0.115\textwidth}p{0.115\textwidth}}
\toprule
 &  Metric (repetitions) & no feedback (\GEN) & w/ feedback (\CORR) \\
\midrule
\atomic & per graph  & 2.05   &  \textbf{1.26} \\\cline{2-4}
                           & \% graphs  &  72    &  \textbf{48} \\
\midrule
\snli & per graph   & 2.09   &  \textbf{1.18} \\\cline{2-4}
                           & \% graphs &  73     &   \textbf{46} \\
\midrule
\social & per graph   & 2.2   & \textbf{1.32 } \\\cline{2-4}
                          & \% graphs & 75    & \textbf{ 49} \\
\midrule
Average & per graph   & 2.11   & \textbf{1.25} \\\cline{2-4}
                          & \% graphs & 73.3    & \textbf{ 47.6} \\
\midrule

\end{tabular}
\caption{\CORR reduces the inconsistencies in the graphs. The number of repetitions on average per graph and percentage of graphs with some repetition, both improve.   
}
\label{tab:rq1-main-defeasible}
\end{table}

\subsection{Does using graphs generated by \CORR help the end task?}
\label{sec:rq2}

We now evaluate the efficacy of the graphs generated by \GEN and corrected by \CORR on the defeasible inference task.
As mentioned in Section~\ref{sec:task-defeasible}, the goal of the defeasible inference task is to classify an update \upd as a strengthener or weakener of a hypothesis \hypo in the context of premise \pre.

Let $M$ be the graph generated by \GEN for the query \phu.
The graph $M$ and the feedback $F$ on $M$ are then supplied to \CORR to obtain $G$.
We overload the notation and use $M$ and $G$ to refer to the nodes of the graphs generated by \GEN and \CORR, respectively.
Thus, given each defeasible query \phu, we obtain $M$: the set of nodes generated \GEN and $G$: the set of nodes generated by \CORR.

Following \citet{rudinger-etal-2020-thinking}, we prefix a given sequence of tokens $T$ with a special beginning-of-sequence (BOS) token.
$T$ is then encoded using RoBERTa-base~\cite{liu2019roberta}\footnote{We use the implementation by~\cite{wolf2019huggingface}}, and
the hidden representation corresponding to BOS is passed to a classifier (single-layer MLP).
We train three classifiers, each following the above-described architecture with different inputs: (i) Baseline: $T = P \Vert H \Vert S$, (ii) \GEN:  $T = P \Vert H \Vert M \Vert S$, and (iii) \CORR: $T = P \Vert H \Vert G \Vert S$. We report the results in Table~\ref{tab:rq2-main-defeasible}, and observe that: (i) Despite the relative simplicity of our approach (concatenating nodes with the query), both \GEN (concatenates noisy graph) and \CORR (concatenates cleaner graph) improve over the baseline. This shows that these explanation structures help enrich the context in the defeasible reasoning task. (ii) \CORR outperforms both the baseline and \GEN, showing that reducing the inconsistencies and repetitions improves end task performance.

\begin{table}[!h]
\centering
\begin{tabular}{lrrr}
\toprule
        
 & Baseline & \GEN & \CORR \\
\midrule
\atomic   & 78.3   & 78.8    & \textbf{79.5} \\
\snli    & 81.6   & 82.1    & \textbf{83.1} \\
\social  & 86.2   & 86.7    & \textbf{87.2}  \\
\midrule
average  & 82.03   & 82.53    & \textbf{83.26*}  \\
\bottomrule
\end{tabular}
\caption{Results on Defeasible inference without using graphs (Baseline~\cite{rudinger-etal-2020-thinking}), using graphs generated by \GEN, and graphs corrected with feedback by \CORR. {*} indicates statistical significance}
\label{tab:rq2-main-defeasible}
\end{table}

\section{Discussion and Conclusion} 
We present \ours, a system that improves the explanation structure (graphs) generated by a model without requiring expensive human-annotated feedback.
Our approach generates graphs that have 40\% fewer inconsistencies as compared with the off-the-shelf system.
Further, simply appending the corrected explanation structures to the output leads to a gain of 1.2 points on accuracy on defeasible reasoning across all three domains.

This work paves a new path towards exciting future research direction of constantly improving explainable NLP models by applying human feedback.

\section*{Acknowledgments}
This material is partly based on research sponsored in part by the Air Force Research Laboratory under agreement number FA8750-19-2-0200. 
The U.S. Government is authorized to reproduce and distribute reprints for Governmental purposes notwithstanding any copyright notation thereon. 
The views and conclusions contained herein are those of the authors and should not be interpreted as necessarily representing the official policies or endorsements, either expressed or implied, of the Air Force Research Laboratory or the U.S. Government.
We would like to thank Google for providing the TPU machines for conducting experiments.

\bibliographystyle{acl_natbib}
\bibliography{acl2021}
\newpage

\appendix
\appendix

\section{Appendix}
\label{sec:appendix}

\subsection{Examples of errors in the explanation structures generated by \GEN}
\begin{figure*}
\centering
\begin{minipage}{.5\textwidth}
  \centering
  \includegraphics[width=.8\linewidth]{imgs/snli_bad.pdf}
\end{minipage}%
\begin{minipage}{.5\textwidth}
  \centering
  \includegraphics[width=.8\linewidth]{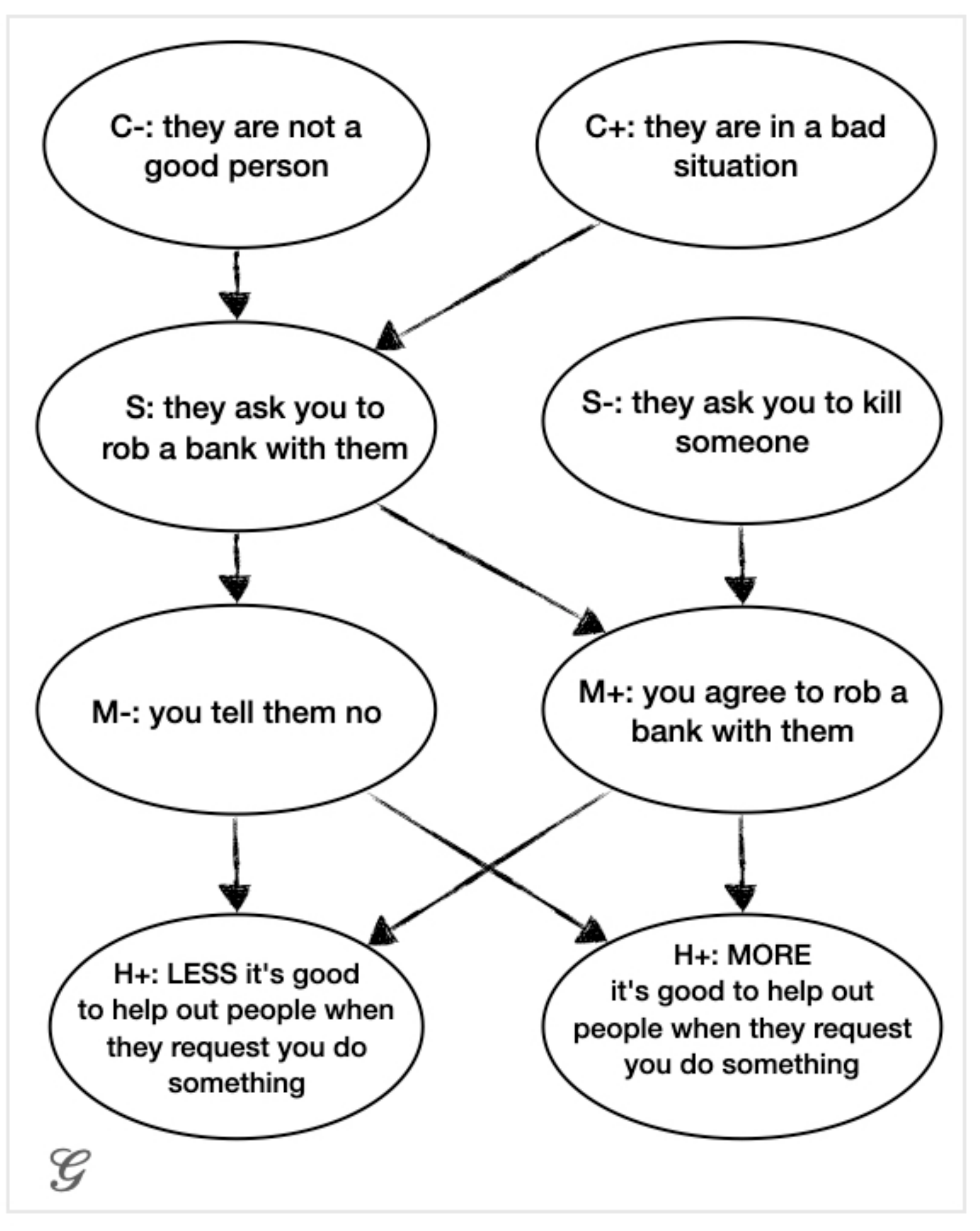}
\end{minipage}
  \captionof{figure}{Incorrect graph generated by \GEN (left) and fixed by \CORR (right) for Defeasible-SNLI dataset. The feedback is `C-, C+ are overlapping, and S, S- are overlapping.'}
  \label{fig:snliexample}
\end{figure*}

\begin{figure*}
\centering
\begin{minipage}{.5\textwidth}
  \centering
  \includegraphics[width=.8\linewidth]{imgs/social_bad.pdf}
\end{minipage}%
\begin{minipage}{.5\textwidth}
  \centering
  \includegraphics[width=.8\linewidth]{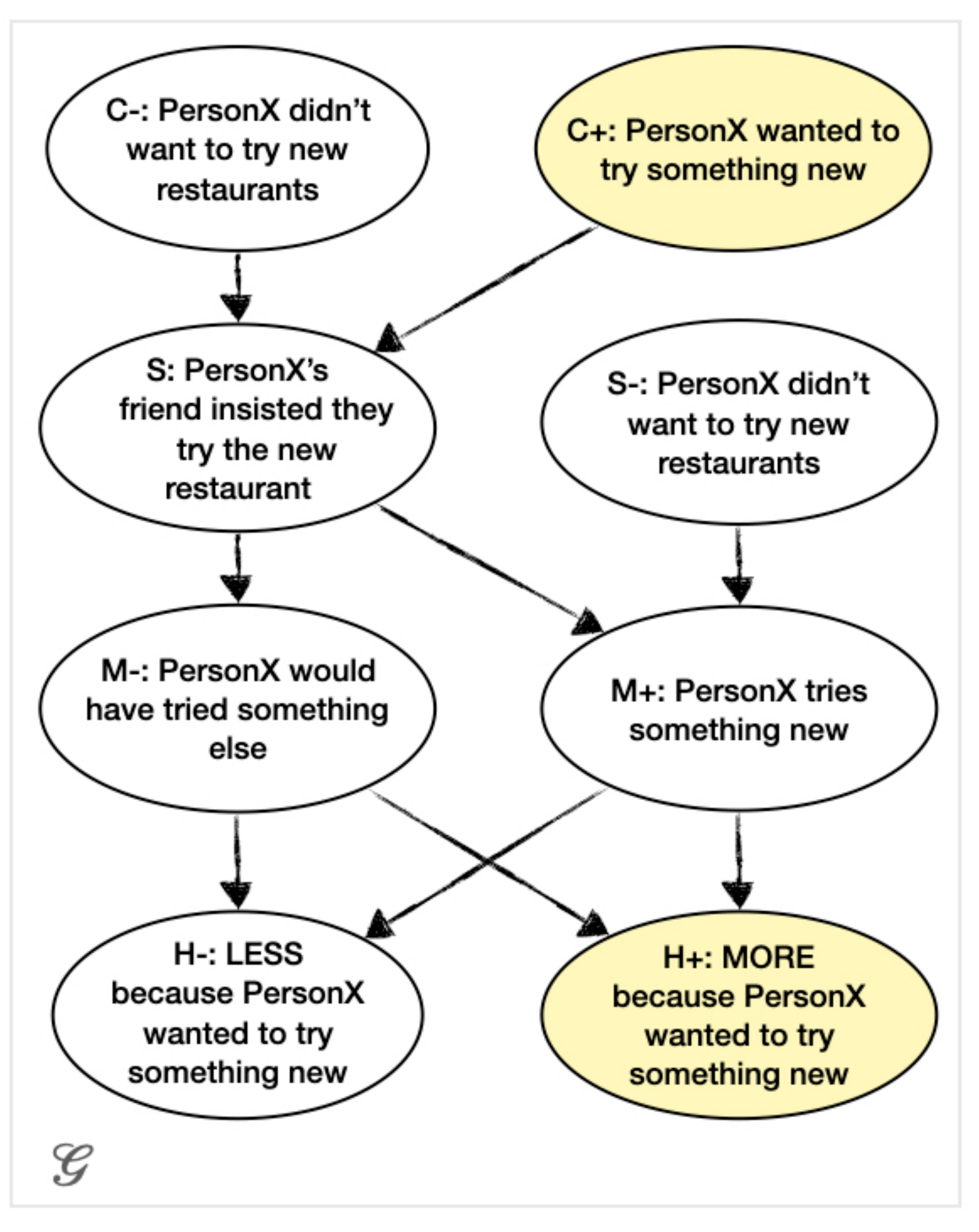}
\end{minipage}
  \captionof{figure}{Incorrect graph generated by \GEN (left) and fixed by \CORR (right) for Defeasible-SOCIAL dataset. The feedback is `C-, C+,S,S- are overlapping, and M-, M+, H+ are overlapping.'}
  \label{fig:socialexample}
\end{figure*}

\begin{figure*}
\centering
\begin{minipage}{.5\textwidth}
  \centering
  \includegraphics[width=.8\linewidth]{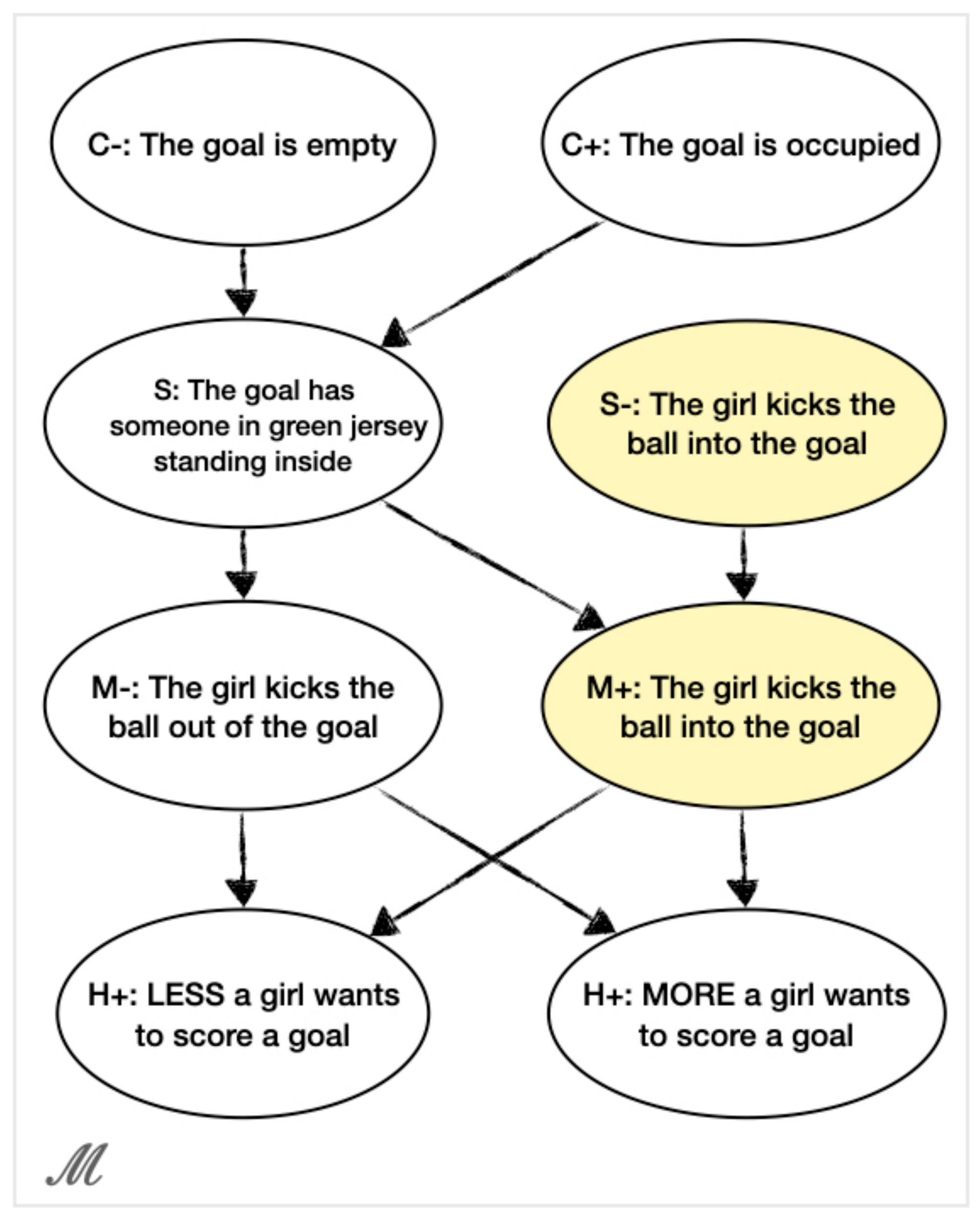}
\end{minipage}%
\begin{minipage}{.5\textwidth}
  \centering
  \includegraphics[width=.8\linewidth]{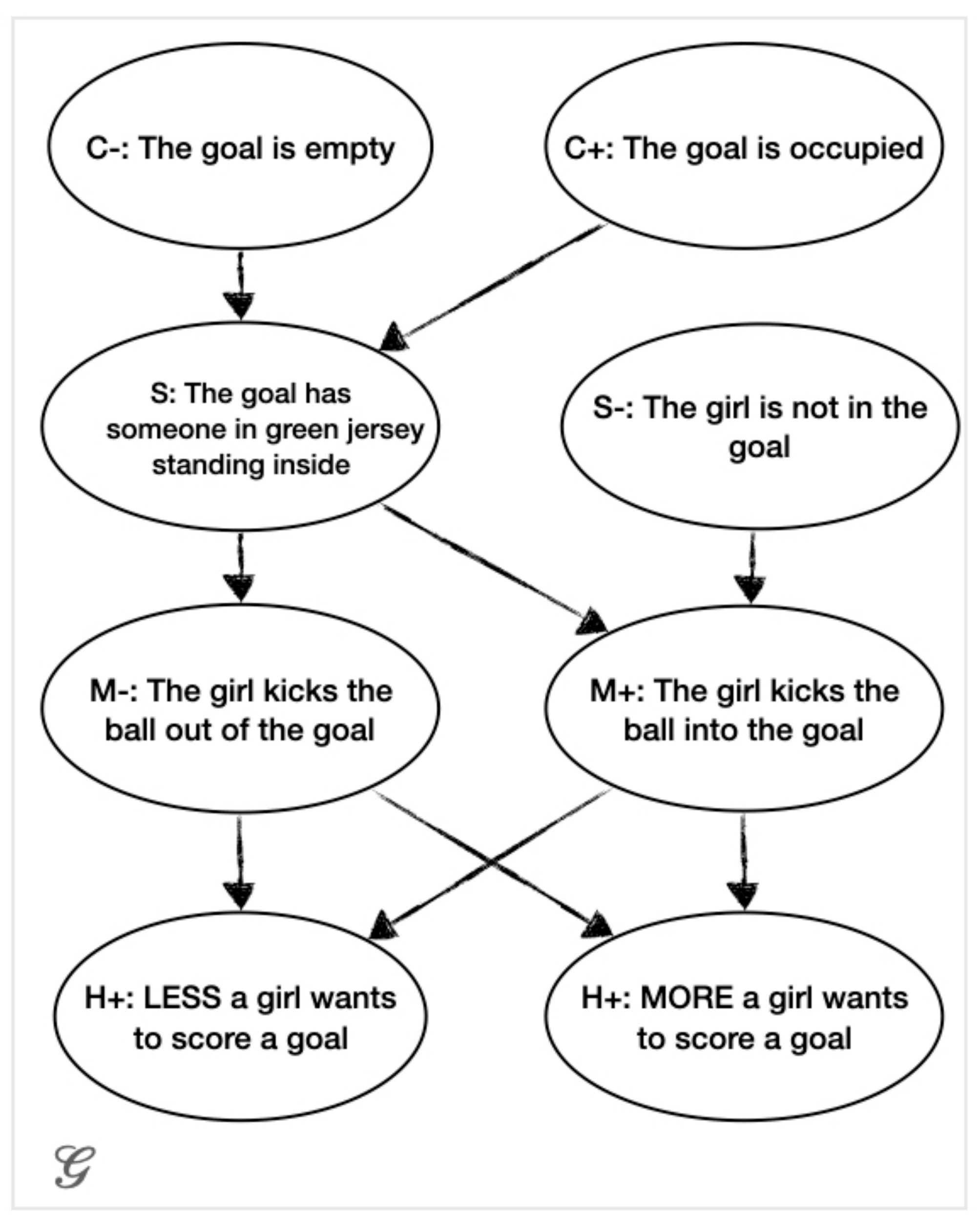}
\end{minipage}
  \captionof{figure}{Incorrect graph generated by \GEN (left) and fixed by \CORR (right) for Defeasible-ATOMIC dataset. The feedback is `S-, M+ are overlapping.;}
  \label{fig:atomicexample}
\end{figure*}

\squishlist
\item Figure \ref{fig:snliexample} shows an example of incorrect graph generated for Defeasible SNLI data.
\item Figure \ref{fig:socialexample} shows an example of incorrect graph generated for Defeasible Social data.
\item Figure \ref{fig:atomicexample} shows an example of incorrect graph generated for Defeasible ATOMIC data.
\squishend

\subsection{Reproducibility}
\subsubsection{\GEN, \GENS, \CORR}
T5-11B models has 11B parameters with 24-layers, 1024-hidden-state, 65,536 feed-forward hidden-state, 128 attention heads.
We use TPU (v3-8) on Google cloud platform. 
It takes 3 hours in average to train \GEN and \GENS, and 4 hours to train \CORR.

\subsubsection{Classifier for defeasible tasks}
We build on the implementation by~\citet{wolf2019huggingface}, using the default hyperparameters.
For optimization, we use AdamW~\cite{loshchilov2017decoupled} with a learning rate of 2e-5, a batch size of 16, and a linear rate scheduler with warm up for the first 3 (10\%) of the epochs.
We use accumulate gradients for two batches, and clip gradients at 1.
We also experimented with a block size of 300 and a batch size of 2.
All of our experiments were done on a single Nvidia GeForce \textsc{rtx} 2080 Ti.

\end{document}